\documentclass[10pt,twocolumn,letterpaper]{article}

\usepackage{cvpr}              

\usepackage{graphicx}
\usepackage{marvosym}
\usepackage{amsmath}
\usepackage{amssymb}
\usepackage{booktabs}
\usepackage{threeparttable}
\usepackage{bbm}
\usepackage{float}
\usepackage{multirow}
\usepackage[accsupp]{axessibility}
\usepackage[pagebackref,breaklinks,colorlinks]{hyperref}

\usepackage[capitalize]{cleveref}
\crefname{section}{Sec.}{Secs.}
\Crefname{section}{Section}{Sections}
\Crefname{table}{Table}{Tables}
\crefname{table}{Tab.}{Tabs.}

\begin{document}
\title{Submission to Generic Event Boundary Detection Challenge@CVPR 2022:\\
Local Context Modeling and Global Boundary Decoding Approach}

\author{Jiaqi Tang$^1$ \qquad Zhaoyang Liu$^{2,3}$ \qquad Jing Tan$^1$ \qquad Chen Qian$^{2,3}$ \qquad Wayne Wu$^{2,3}$ \qquad Limin Wang$^{1}$\\
$^1$State Key Laboratory for Novel Software Technology, Nanjing University, China\\
$^2$SenseTime Research \qquad $^3$Shanghai AI Laboratory\\
{\tt\small jqtang@smail.nju.edu.cn, zyliumy@gmail.com, jtan@smail.nju.edu.cn} \\  {\tt\small \{qianchen, wuwenyan\}@sensetime.com, lmwang@nju.edu.cn}\\
}
\maketitle
\begin{abstract}
Generic event boundary detection (GEBD) is an important yet challenging task in video understanding, which aims at detecting the moments where humans naturally perceive event boundaries. In this paper, we present a local context modeling and global boundary decoding approach for GEBD task. Local context modeling sub-network is proposed to perceive diverse patterns of generic event boundaries, and it generates powerful video representations and reliable boundary confidence. Based on them, global boundary decoding sub-network is exploited to decode event boundaries from a global view. Our proposed method achieves 85.13\% F1-score on Kinetics-GEBD testing set, which achieves a more than 22\% F1-score boost compared to the baseline method. The code is available at \url{https://github.com/JackyTown/GEBD_Challenge_CVPR2022}.
\end{abstract}

\section{Introduction}
\label{sec:intro}
Cognitive science~\cite{tversky2013event} suggests that humans naturally divide a video into meaningful units by perceiving event boundaries. To this end, a task termed as \textbf{Generic Event Boundary Detection}~\cite{Shou_2021_ICCV} (GEBD) is recently proposed to localize the generic event boundaries in videos, which is expected to facilitate the development of long-form video understanding.

We leverage DDM-Net~\cite{ddm22cvpr} to model local temporal contexts of generic event boundaries, the motivation is three-fold.
The \emph{primary} challenge in GEBD task is to model diverse patterns of generic event boundaries.
Spatial diversity is dominantly characterized by the change of appearance, which normally comprises low-level changes (\eg, change in environment or brightness) and high-level changes (\eg, action changes). Temporal diversity is mainly relevant to actions. Notably, different actions usually exhibit inconsistent speed and duration. To this end, we exploit a feature bank to store multi-level features of space and time. \emph{Second}, previous motion representation, such as optical flow and RGB differences, only focus on local motion clues between two consecutive frames. In contrast, we present dense difference maps for long-range and holistic temporal context modeling. \emph{Third}, previous two-stream methods are not sufficient to aggregate the appearance and motion clues. Naive fusion methods like linear fusion and concatenation lack interaction between modalities. Therefore, we employ progressive attention to make use of correlations between two modalities.

Furthermore, we contend that video-level global temporal context modeling is vital for GEBD. Viewing videos as temporal sequences and employing Transformer architecture to model global one-dimensional dependencies boosts localization performance. We exploit a direct event boundary detection framework with Transformers, termed as RTD-Net~\cite{Tan_2021_ICCV}. This direct event boundary detection with parallel decoding allows us to better capture inter-boundary relationships from a global view, thus resulting in more complete and precise localization results.

As a result, our approach outperforms the existing baseline method PC~\cite{Shou_2021_ICCV} by a large margin on all evaluation metrics. Particularly, our approach obtains 85.13\% F1-score@0.05 on Kinetics-GEBD testing set, with a significant boost of more than 22 percent.

\begin{figure*}[t]
 \centering
 \subfloat[Overview of local context modeling sub-network]{\includegraphics[width=6.6in]{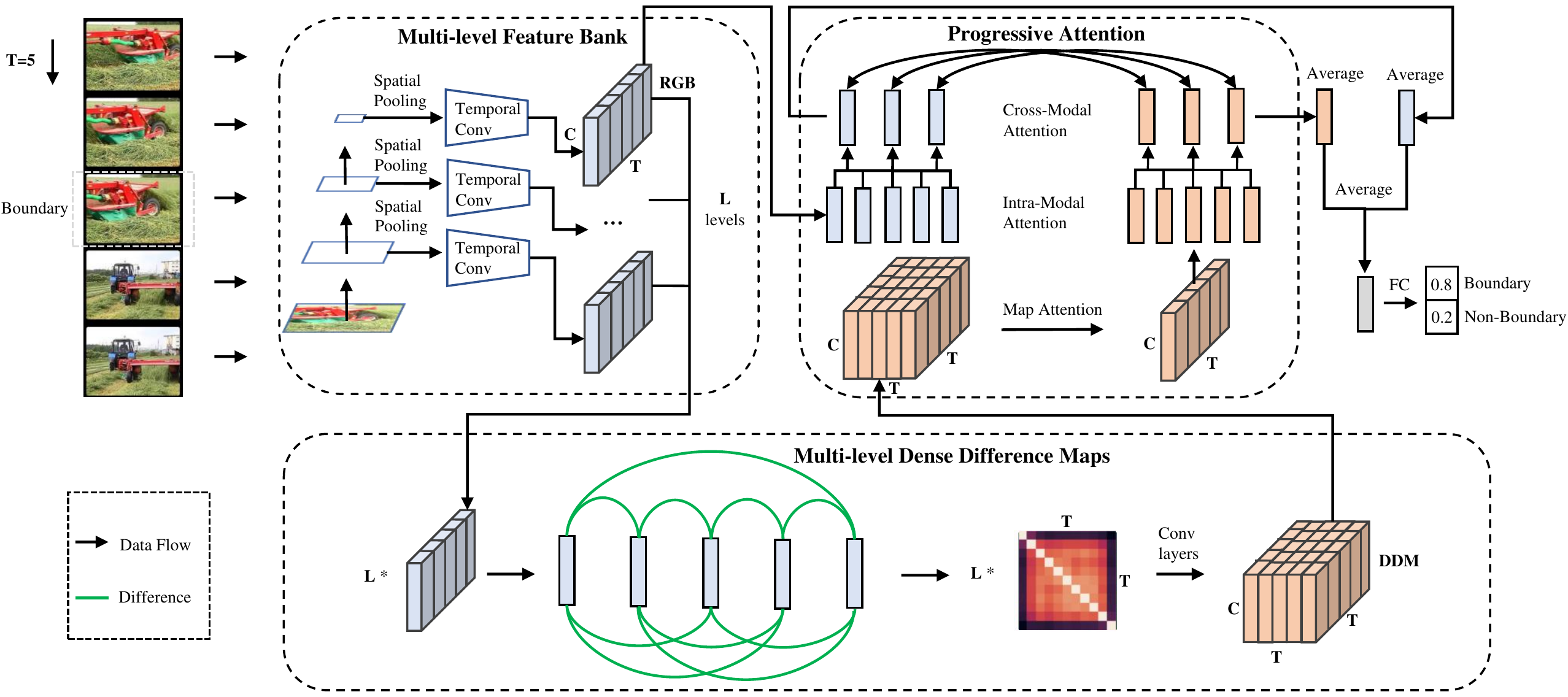}} 
 \label{s}\\
 \subfloat[Overview of global boundary decoding sub-network]{\includegraphics[width=6.6in]{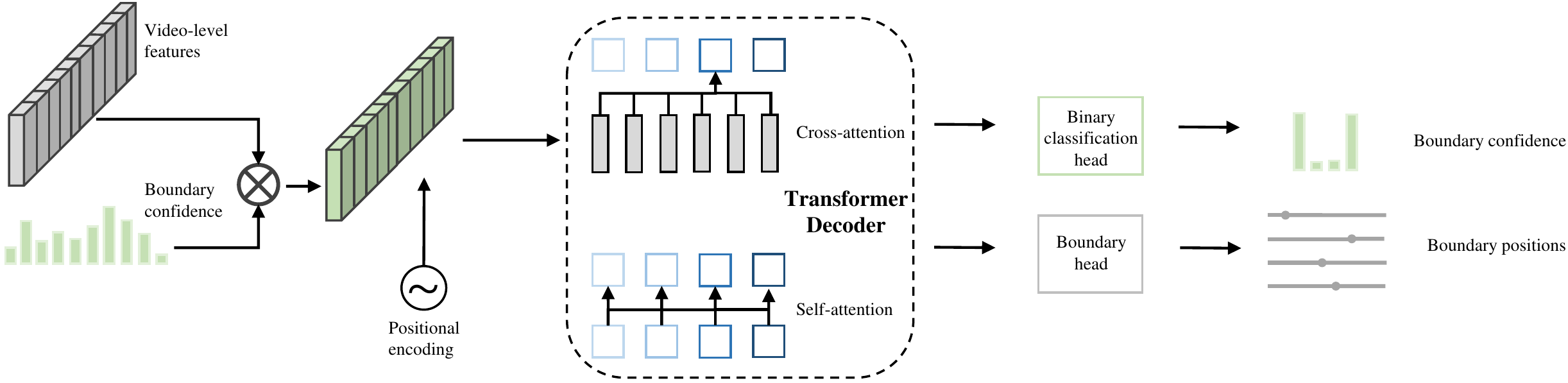}}
 \caption{{\bf Overview of our approach.} (a) is DDM-Net~\cite{ddm22cvpr} for local context modeling, (b) is RTD-Net~\cite{Tan_2021_ICCV} for global boundary decoding.}
 \vspace{-6mm}
 \label{fig:overview}
\end{figure*}
\section{Related Work}
\noindent\textbf{Temporal Action Detection.} The goal of temporal action detection is to localize actions in untrimmed videos flexibly and precisely. Anchor-based methods~\cite{DBLP:conf/cvpr/HeilbronNG16, DBLP:conf/iccv/GaoYSCN17, DBLP:conf/iccv/XuDS17, DBLP:conf/cvpr/ChaoVSRDS18} retrieved proposals based on multi-scale and dense anchors, which are inflexible and cannot cover all action instances. Boundary-based methods~\cite{DBLP:conf/iccv/ZhaoXWWTL17, DBLP:conf/eccv/LinZSWY18, DBLP:conf/iccv/LinLLDW19} first evaluated the confidence of starting and ending points and then matched them to form proposal candidates. However, they generated results only based on local information and were easily affected by noise. In contrast, RTD-Net~\cite{Tan_2021_ICCV} makes predictions based on the whole feature sequence and fully leverages the global temporal context. 

\noindent\textbf{Generic Event Boundary Detection.}  Generic event boundary detection~\cite{Shou_2021_ICCV} (GEBD) is a novel task for detecting taxonomy-free moments that segment a video into trunks, where humans naturally perceive event boundaries. GEBD is a generic detection task in video understanding, where generic event boundaries include action, shot and anomaly boundaries. To address the diversity and complicated semantics of generic event boundaries, DDM-Net~\cite{ddm22cvpr} improves the boundary discrimination via progressively attending to multi-level dense difference maps.

\section{Approach}
Our approach is mainly composed of two sub-networks. The first is DDM-Net~\cite{ddm22cvpr} for local context modeling, and the second is RTD-Net~\cite{Tan_2021_ICCV} for global boundary decoding. As illustrated in Figure~\ref{fig:overview}, our approach mainly refers to four stages. Firstly, DDM-Net is trained to distinguish boundaries and non-boundaries. Secondly, accurate boundary confidence and powerful representations of training, validation and testing set are generated via the inference of DDM-Net. Thirdly, based on features of the second stage, RTD-Net is trained to predict event boundaries via a set of boundary queries. Finally, we exploit RTD-Net to decode boundary predictions of the testing set.
\subsection{Local Context Modeling}
DDM-Net~\cite{ddm22cvpr} is leveraged to model local temporal contexts, which consists of three stages: Multi-Level Feature Bank construction, Dense Difference Map calculation and Progressive Attention.

\noindent\textbf{Multi-Level Feature Bank.} Patterns of different event boundaries vary considerably in space and time. To address spatial diversity, we perform average spatial pooling on $m$ layers of backbone features with different spatial resolutions and get $m$ feature sequences of different semantic levels. To handle temporal diversity, for each feature sequence we exploit temporal convolutions to get $n$ feature sequences with different temporal receptive fields. As a result, there are $m\times n=L$ levels of features in total.

\noindent\textbf{Dense Difference Maps.} Sequential optical flow or RGB differences can only reflect local motion cues between two consecutive frames, which is insufficient for diverse and complicated scenarios in GEBD task. Instead, given a feature sequence of $T$ frames, we calculate the feature difference of each frame pair and construct a $T\times T$ map. Compared with the sparse motion sequence of length $T-1$, $T\times T$ pairs of feature differences provide denser temporal cues and more holistic temporal contexts around the center frame, enabling our method to
better perceive temporal variations and distinguish boundaries and non-boundaries.

\noindent\textbf{Progressive Attention.} As naive fusion methods lack interaction between modalities, they are less effective to exploit feature complementarity of appearance and motion features. To this end, we employ Progressive Attention to perform cross-modal interaction. Map Attention aligns DDM $M\in \mathbb{R}^{C\times T\times T}$ with RGB features $A\in \mathbb{R}^{C\times T}$. Intra-Modal Attention adaptively aggregates and enhances key representations with $\omega$ learnable queries. Cross-Modal Attention jointly learns features across modalities through co-attention transformers.

\subsection{Global Boundary Decoding}
We exploit RTD-Net~\cite{Tan_2021_ICCV} to decode event boundaries from a global view. First, we enhance DDM-Net representations with DDM-Net boundary scores. Then, the transformer decoder uses a set of learnable queries to attend to the boundary-attentive representations. Eventually, boundary head and binary classification head transform query embedding to final boundary predictions.

\noindent\textbf{Boundary-Attentive Representations.} To alleviate the issue of feature slowness~\cite{DBLP:journals/pami/ZhangT12}, we propose the boundary-attentive module to explicitly enhance short-term features with discriminative boundary information. Specifically, we multiply the original features with their boundary confidence, which is estimated with DDM-Net~\cite{ddm22cvpr}.

\noindent\textbf{Transformer Decoder.} We use the vanilla Transformer decoder to directly output event boundaries. The decoder takes a set of boundary queries and boundary-attentive representations as input, and outputs event boundary embedding for each query via stacked multi-head self-attention and cross-attention blocks. Self-attention layers model the temporal dependencies between event boundaries and refine the corresponding query embedding. In cross-attention layers, boundary queries attend to all time steps and aggregate change information at high activation into each boundary embedding. During the training procedure, this decoder collaborates with a Hungarian matcher to align positive boundaries with ground truths and the whole pipeline is trained with a set prediction loss.

\noindent\textbf{Feed Forward Network (FFN).} 
RTD-Net generates boundary predictions by designing two feed forward networks (FFNs) as detection heads. Boundary head decodes temporal locations of event boundaries. Binary classification head predicts boundary confidence $p_{bc}$ of each sampled frame. In inference, if the boundary confidence of a frame is greater than a threshold $\theta$, it is considered an event boundary.

\section{Experiments}
\subsection{Dataset and Setup}
\noindent\textbf{Kinetics-GEBD.} Kinetics-GEBD dataset~\cite{Shou_2021_ICCV} consists of nearly 60,000 videos selected from Kinetics-400. The ratio of training, validation and testing set is nearly 1:1:1. We use both Kinetics-GEBD training and validation set to train local context modeling sub-network. For global boundary decoding sub-network, we randomly sample 1,000 videos from 
Kinetics-GEBD validation set to construct a local validation
set and use all the other samples for training.

\noindent\textbf{Evaluation Protocol.}
To evaluate the results of generic event boundary detection task, we calculate F1-score and use the Relative Distance (Rel.Dis.) measurement~\cite{Shou_2021_ICCV}. Rel.Dis. is the relative distance between predictions and ground truths, divided by the length of the corresponding video. Given a threshold, a prediction is determined to be true if Rel.Dis. is smaller than or equal to the threshold, otherwise false. 

\subsection{Implementation Details}
\noindent\textbf{Local Context Modeling Sub-network.} The input of DDM-Net~\cite{ddm22cvpr} is a local clip centered on the current frame, and the output is boundary confidence of the frame and features of the clip (local contexts of the current frame). We select one frame out of every 3 consecutive frames, namely the stride of boundary evaluation is 3. To predict the boundary confidence of the current frame, we take a $T\times s$ ($T=2 \times w + 1$) clip as the input, where $w$ is 16 and $s$ is 2. In practice, DDM-Net is built on IG-65M pretrained CSN backbone and trained end-to-end. $m$ and $n$ of multi-level feature bank are set to 3. $\omega$ of progressive attention is set to 5. To train DDM-Net, we employ Adam as the optimizer. The batch size is set to 16 and the learning rate is set to 1e-5. 

\noindent\textbf{Global Boundary Decoding Sub-network.}
We get video-level representations via stacking outputs of DDM-Net (features of local clips). RTD-Net~\cite{Tan_2021_ICCV} takes the representation of a video as input and outputs final boundary predictions of the video. On Kinetics-GEBD, we perform boundary decoding in a sliding window manner and the length of each sliding window is set to 100. We re-scale short feature sequences to 100 via padding features of the last clip. To train RTD-Net from scratch, we use AdamW for optimization. The batch size is set to 32 and the learning rate is set to 1e-4. The number of boundary queries is 10 and the threshold $\theta$ is set to 0.87.

\subsection{Main Results}
As demonstrated in Table~\ref{table:val}, we fairly compare our DDM-Net~\cite{ddm22cvpr} with PC~\cite{Shou_2021_ICCV} on the validation set of Kinetics-GEBD. DDM-Net outperforms PC by a large margin. For detailed ablation studies of DDM-Net, please refer to ~\cite{ddm22cvpr}. When DDM-Net is combined with IG-65M~\cite{DBLP:conf/cvpr/GhadiyaramTM19} pretrained CSN~\cite{DBLP:conf/iccv/TranWFT19}, it can further increase the F1-score by nearly 5 percent, from 76.4\% to 81.3\%. Furthermore, if we use both Kinetics-GEBD training and validation set for training DDM-Net and RTD-Net~\cite{Tan_2021_ICCV}, we can achieve 86.08\% F1-score on our local validation set with 1,000 videos.

As a result, our submission achieves 85.13\% F1-score on Kinetics-GEBD testing set, as shown in Table~\ref{table:test}. It is worth noting that the result is obtained with only DDM-Net combined with IG-65M pretrained CSN backbone and RTD-Net. 

\begin{table}[!h]
\begin{center}
\caption{Main results on Kinetics-GEBD validation set.}
\vspace{-3mm}
\scalebox{0.82}{
\begin{tabular}{c|cccc}
\toprule[1pt]
Model & F1-score & Precision & Recall \\ 
\hline
PC (R50) & 0.625 & 0.624 & 0.626 \\
DDM-Net (R50) & 0.764 & 0.732 &	0.800 \\
DDM-Net (CSN) & 0.813 & 0.802 & 0.824 \\
DDM-Net (CSN) + RTD-Net* & \textbf{0.861} & \textbf{0.824} &	\textbf{0.901} \\
\bottomrule[1pt]
\end{tabular}
}
\begin{tablenotes}
\footnotesize
\item * results are reported based on our local validation set.
\end{tablenotes}
\vspace{-4mm}
\label{table:val}
\end{center}
\end{table}
\vspace{-2mm}

\begin{table}[!h]
\begin{center}
\caption{Main results on Kinetics-GEBD testing set.}
\vspace{-3mm}
\scalebox{0.82}{
\begin{tabular}{c|cccc}
\toprule[1pt]
Model & F1-score & Precision & Recall \\ 
\hline
PC (R50) & 0.625 & 0.624 & 0.626 \\
DDM-Net (CSN) & 0.837 & 0.837 &	0.837 \\
DDM-Net (CSN) + RTD-Net & \textbf{0.851} & \textbf{0.84} &	\textbf{0.86} \\
\bottomrule[1pt]
\end{tabular}
}
\vspace{-4mm}
\label{table:test}
\end{center}
\end{table}

\section{Conclusion}
In this paper, we introduce our local context modeling and global boundary decoding approach for GEBD Challenge@CVPR 2022, which consists of DDM-Net for local context modeling and RTD-Net for global boundary decoding. Compared with the given baseline PC, our approach achieves a remarkable performance gain on Kinetics-GEBD validation and testing set.

{\small
\bibliographystyle{ieee_fullname}
\bibliography{egbib}
}

\end{document}